\documentclass[conference]{IEEEtran}
\IEEEoverridecommandlockouts
\usepackage{multirow}

\usepackage[noadjust]{cite}

\usepackage{amsmath,amssymb,amsfonts}
\usepackage{algorithmic}
\usepackage{graphicx}
\usepackage{amsmath}
\usepackage{textcomp}
\usepackage{xcolor}
\usepackage{comment}
\renewcommand{\baselinestretch}{0.93}
\def\BibTeX{{\rm B\kern-.05em{\sc i\kern-.025em b}\kern-.08em
    T\kern-.1667em\lower.7ex\hbox{E}\kern-.125emX}}

\usepackage{booktabs}

\begin{document}


\title{\LARGE Code-Bridged Classifier (CBC): A Low or Negative Overhead Defense for Making a CNN Classifier Robust Against Adversarial Attacks 
}

\author{\IEEEauthorblockN{Farnaz Behnia, Ali Mirzaeian, Mohammad Sabokrou, Saj Manoj, Tinoosh Mohsenin, \\ Khaled N. Khasawneh, Liang Zhao, Houman Homayoun, Avesta Sasan }
fbehnia@gmu.edu, amirzaei@gmu.edu, sabokro@ipm.ir, spudukot@gmu.edu, tinoosh@umbc.edu, \\ kkhasawn@gmu.edu,  
lzhao9@gmu.edu, hhomayoun@ucdavis.edu, asasan@gmu.edu}

\maketitle
\begin{abstract}


In this paper, we propose Code-Bridged Classifier (CBC), a framework for making a Convolutional Neural Network (CNNs) robust against adversarial attacks without increasing or even by decreasing the overall models' computational complexity. More specifically, we propose a stacked encoder-convolutional model, in which the input image is first encoded by the encoder module of a denoising auto-encoder, and then the resulting latent representation (without being decoded) is fed to a reduced complexity CNN for image classification. We illustrate that this network not only is more robust to adversarial examples but also has a significantly lower computational complexity when compared to the prior art defenses. 

\end{abstract}
\section{Introduction}\label{sec:introduction}
Deep learning is the foundation for many of today's applications, such as computer vision, natural language processing, and speech recognition. After AlexNet \cite{alexnet} made a breakthrough in 2012 by significantly outperforming other object detection solutions, and winning the ISLVRC competition \cite{islvrc}, CNNs gained a well-deserved popularity for computer vision applications. This energized the research community to architect models capable of achieving higher accuracy (that led to development of many higher accuracy models including GoogleNet \cite{googlenet} and ResNet \cite{resnet}), increased the demand and research for hardware platforms capable of fast execution of these models \cite{NcdNpeAmirzaei,nesta}, and created a demand for lower complexity models \cite{icnn,icnntecs, exploit} capable of reaching high levels of accuracy. 

Even though the evolution in their model structure and the improvement in their accuracy have been very promising in recent years, it is illustrated that convolutional neural networks are prone to adversarial attacks through simple perturbation of their input images \cite{fgsm ,bim ,mim ,deepfool}. The algorithms proposed by \cite{fgsm ,bim , mim, deepfool} have demonstrated how easily the normal images can be perturbed with adding a small noise in order to fool neural networks. The main idea is to add a noise vector containing small values to the original image in the opposite or same direction of the gradient calculated by the target network to produce adversarial samples \cite{fgsm , bim}.

The wide-spread adoption of CNNs in various applications and their unresolved vulnerability to adversarial samples has raised many safety and security concerns and has motivated a new wave of deep learning research. To defend against adversarial attacks, the concept of adversarial training was proposed in \cite{fgsm} and was further refined and explored in \cite{bim , mim}. Adversarial training is a data augmentation technique in which by generating a large number of adversarial samples and including them with correct labels in the training set, the robustness of network against adversarial attacks improves.  Training an adversarial classifier to determine if the input is normal or adversarial and using autoencoder (AE) to remove the input image noise before classification are some of the other approaches taken by \cite{fgsm} and \cite{intriguing}. Finally, \cite{distillation} utilizes distillation as a defense method against adversarial attacks in which a network with a similar size to the original network is trained in a way that it hides the gradients between the softmax layer and its predecessor.


In this work, we combine denoising and classification into a single solution and propose the code-bridged classifier (CBC). We illustrate that CBC is 1) more robust against adversarial attacks compared to a similar CNN solution that is protected by a denoising AE, and has substantially less computational complexity compared to such models.\par



\section{Background and Related Work}\label{sec:background}
The vulnerability of deep neural networks to adversarial examples was first investigated in \cite{intriguing}. Since this early work, many new algorithms for generating adversarial examples, and a verity of solutions for defending against these attacks are proposed. Following is a summary of the attack and defense models related to our proposed solution: 

\subsection{Attack Models}
Many effective attacks have been introduced in the literature. Some of the most notable attacks include Fast Gradient Sign Method (fgsm) \cite{fgsm}, Basic Iterative Method \cite{bim}, Momentum Iterative Method \cite{mim}, DeepFool \cite{deepfool} and Carlini Wagner~\cite{cw}, the description of each method are as follows.

\subsubsection {FGSM attack} 
\begin{figure}[t]
  \centering
    \includegraphics[width=0.85\columnwidth]{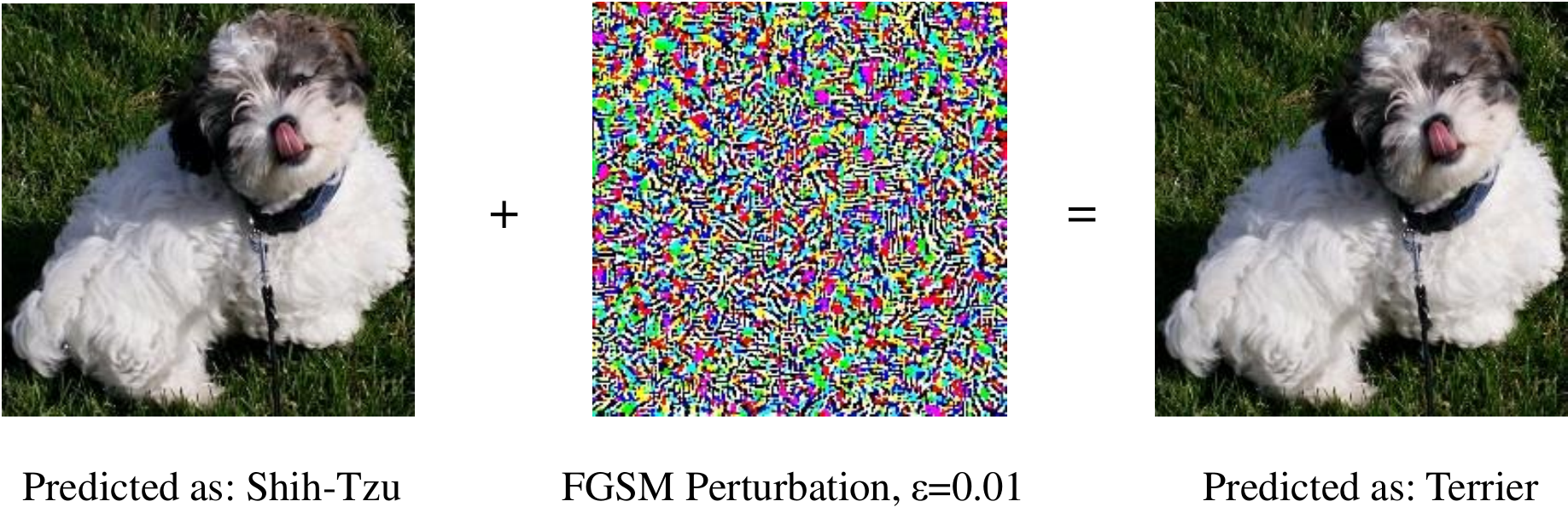}
  \caption{The FGSM attack is used to add adversarial noise to the original image. The adversarial perturbation remains imperceptible to the human eyes but causes the neural network to misclassify the input image.}
  \label{fig:fgsm}
\end{figure}

in \cite{fgsm}, a simple method is suggested to add a small perturbation to the input to make an adversarial image. The adversarial image is obtained by:
\begin{equation}\label{eq_fgsm}
    x'= x + \epsilon sign(\nabla_x J(\theta, x, y))
\end{equation}
in which $x$ is the input image, $y$ is the correct label, $\theta$ is the network parameters, and $J$ is the loss function. $\epsilon$ defines the magnitude of the noise. The larger the $\epsilon$, the larger the possibility of misclassification. Figure \ref{fig:fgsm} illustrates how such adversarial perturbation can change the classifier's prediction.

\subsubsection {Basic Iterative Method (BIM) attack \cite{bim}}  
Also known as Iterative-FGSM attack, BIM attack is iterating over the FGSM attack, increasing the effectiveness of the attack. The BIM attack can be expressed as:
\begin{equation}\label{eq-bim}
x'_0=x ,~x'_n = x'_{n-1} + \epsilon sign(\nabla_x J(\theta, x_{n-1}, y))
\end{equation}

\subsubsection {Momentum Iterative attack \cite{mim}} 
In the Momentum Iterative attack, the momentum is also considered when calculating the adversary perturbation, and is expressed as:
  \begin{equation}\label{eq-mim}
  \begin{aligned}
g_0 = 0, ~g_{n}=\mu g_{n-1}\frac{\nabla_x J(\theta, x_{n-1}, y)}{||\nabla_x J(\theta, x_{n-1}, y)||_1}\\
\\
x'_n = x'_{n-1} + \epsilon sign(g_n)
\end{aligned}
\end{equation}
in which $\mu$ is the momentum, and $||\nabla_x J(\theta, x_{n-1}, y)||_1$ is the $L_1$ norm of the gradient.

\subsubsection {Deepfool \cite{deepfool}} 
The Deepfool attack is formulated such that it can find adversarial examples that are more similar to the original ones. It assumes that neural networks are completely linear and classes are distinctively separated by hyper-planes. With these assumptions, it suggests an optimal solution to find adversarial examples. However, because neural networks are nonlinear, the step for finding the solution is repeated. We refer to \cite{deepfool} for details of the algorithm.   

\subsubsection { Carlini \& Wagner (CW) \cite{cw}} 
Finding adversarial examples in the CW attack is an iterative process that is conducted against multiple defense strategies. The CW attack uses Adam optimizer and a specific loss function to find adversarial examples that are less distorted than other attacks. For this reason, the CW attack is much slower. Adversarial examples can be generated by employing $L_0$ , $L_2$  and $L_{\infty}$ norms. The objective function in CW attack consider an auxiliary variable $w$ and is defined as:
\begin{equation}\label{cost-cw}
    \delta_i = \frac{1}{2}(tanh(w_i)+1)-x_i
\end{equation}

Then if we consider the $L_2$ norm, this perturbation is optimized with respect to $w$:
 \begin{equation}\label{cw-opt}
    min_w||\delta||_2^2 + c.f(\delta + x)
\end{equation}
in which function $f$ is defined as follows:
\begin{equation}\label{cw-f}
    f(x') = max(max\{Z(x')_i : i \neq t\} - Z(x')_t ,- \kappa)
\end{equation}

in the above equation, $Z(x')$ is the pre-softmax output for class $i$, the parameter $t$ represents the target class, and $\kappa$ is the parameter for controlling the confidence of misclassification.

\subsection{Transferability of Adversarial Examples}

All previously described attacks are carried out in a white-box setting in which the attacker knows the architecture, hyperparameters, and trained weights of the target classifier as well as the existing defense mechanism (if any).  It is very hard to defend against white-box attacks because the attacker can always use the information she has to produce new and working adversarial inputs. However, adversarial attacks can be considered in two other settings: Gray Box and Black Box attacks. In gray box attacks, the attacker knows the architecture but doesn't have access to the parameters, the defense mechanism. In black-box setting the attacker does not know the architecture, the parameters, and the defense method.

Unfortunately, it has been shown that adversarial examples generalize well across different models. In \cite{intriguing} it was shown that many of the adversarial examples that are generated for (and are misclassified by) the original network are also misclassified on a different network that is trained from scratch with different hyperparameters or using disjoint training sets. \par

The findings of \cite{intriguing} are confirmed by the following works, as in \cite {universal}, universal perturbations are successfully found that not only generalize across images but also generalize across deep neural networks. These perturbations can be added to all images and the generated adversarial example is transferable across different models. The work in \cite{transferability,practical} show that adversarial examples that can cause a model to misclassify, can have the same influence on another model that is trained for the same task. Therefore, an attacker can train her dummy model to generate the same output, craft/generate adversarial images on her model, and rely on the transferability of the adversarial examples, being confident that there is a high chance for the target classifier to be fooled.  We argue that our proposed solution can effectively defend  black-box attacks. \par

\subsection{Defenses}

Several works have investigated defense mechanisms against adversarial attacks. In \cite{fgsm},  adversarial training is proposed to enhance the robustness of the model.  In \cite{comparative,magnet}  autoencoders are employed to remove the adversarial perturbation and reconstruct a clean input. In \cite{distillation} distillation is used to hide the gradients of the network from the attacker. Other approaches are also used as a defense mechanism \cite{gradreg,detection,protecting}. In this section, we explore the ideas for defending against adversarial examples.

 \subsubsection{Adversarial Training}
 
 The basic idea of the adversarial training \cite{fgsm} is to train a robust classifier via adding many adversarial examples (that are generated using different attacks) to the training dataset \cite{mim,deepfool,minmax}. The problem with this approach is that it can only make the network robust against known (and trained for) attacks for generating adversarial examples. It also increases the training time significantly.

\subsubsection{Defensive Distillation}

In \cite{distillation} distilling was originally proposed to train a smaller student model from a larger teacher model with the objective that the smaller network predicts the probability of the bigger network. The distillation technique takes advantage of the fact that a probability vector contains more information than only class labels, hence, it is a more effective mean for training a smaller network. For defensive distillation, the second network is the same size as the first network \cite{distillation}. The main idea is to hide the gradients between the pre-softmax and softmax layers to make the attacker's job more difficult. However, it was illustrated in \cite{cw} that this defense can be beaten by using the pre-softmax layer outputs in the attack algorithm and/or choosing a different loss function.


\subsubsection{Gradient Regularization}
Input gradient regularization was fist introduced by \cite{gradreg2} to improve the generalization of training in neural networks by a double backpropagation method. \cite{distillation} mentions the double backpropagation as a defense and  \cite{gradreg} evaluate the effectiveness of this idea to train a more robust neural network. This approach intends to make sure that if there is a small change input, the change in KL divergence between the predictions and the labels also will be small. However, this approach is sub-optimal because of the blindness of the gradient regulation.

\subsubsection{Adversarial Detection}
Another approach taken to make neural networks more robust is to detect adversarial examples before feeding to the network\cite{detection,detection2}. \cite{detection} tries to find a decision boundary to separate adversarial and clean inputs. \cite{detection2} deploys the fact that the perturbation of pixel values by adversarial attack alters the dependence between pixels. By modeling the differences between adjacent pixels in natural images, deviations due to adversarial attacks can be detected.

\subsubsection{Autoencoders}
\begin{figure}[t]
  \centering
  \includegraphics[width=\columnwidth]{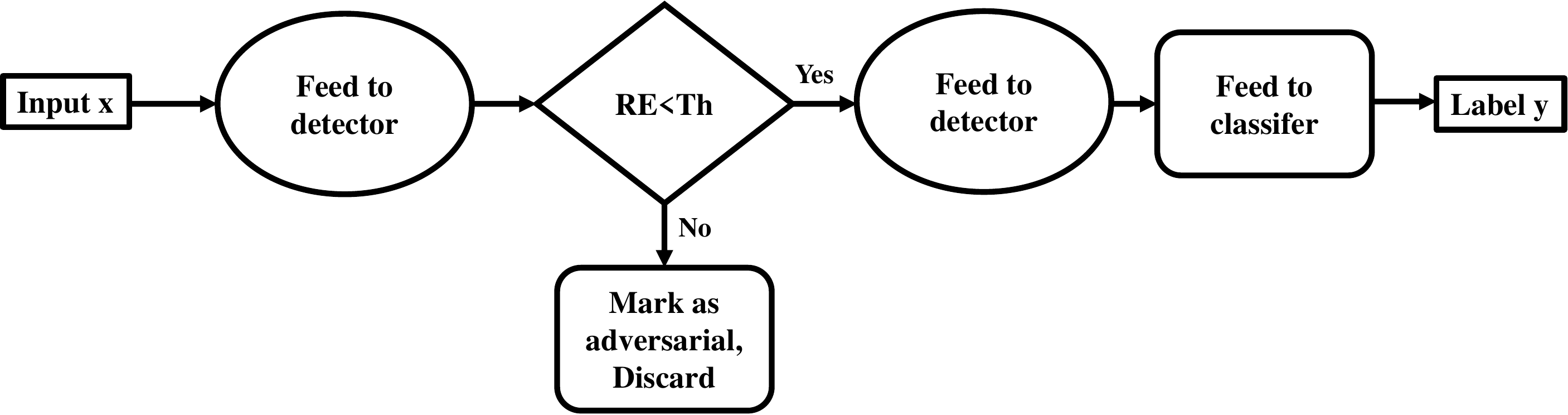}
  \caption {Magnet defense in \cite{magnet} is a two stage defense: the first stage tries to detect the adversarial examples. The images that pass the first stage are denoised using an AutoEncoder in the second stage and fed to the classifier.}
 \label{fig:magnet}
\end{figure}

\cite{comparative} analyzes the use of normal and denoising autoencoders as a defense method. Autoencoders are neural networks that code the input and then try to reconstruct the original image as their output. \cite{magnet}, as illustrated in Fig. \ref{fig:magnet}, uses a two-level module and uses autoencoders to detect and reform adversarial images before feeding to the target classifier. However, this method may change the clean images and also add a computational overhead to the whole defense-classifier module. To improve the method introduced in \cite{magnet},  \cite{sabokrou2019self}  presents an efficient  auto-encoder with a new loss function which is learned to preserve the local neighborhood structure on the data manifold. 

\section{Problem statement}\label{sec:problem}

\begin{figure}[t]
  \centering
  \includegraphics[width=0.99\columnwidth]{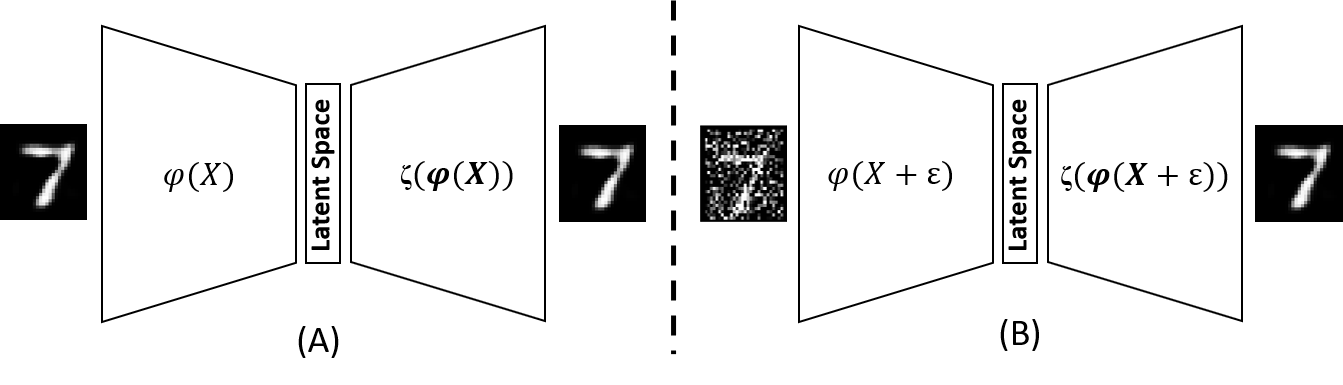}
  \caption {An abstract view of a) a typical Auto-Encoder, and b) a Denoising Autoencoders. Two major components of both structures are 1) Encoder, $\varphi(.)$, which extracts the latent space of sample inputs 2) Decoder, $\zeta(.)$, which reconstructs sample inputs from the latent space.}
 \label{ae}
\end{figure}

An abstract view of a typical Auto-Encoder (AE) and Denoising Auto-Encoder (DAE) is depicted in Fig. \ref{ae}. An AE is comprised of two main components:  1) The \textit{encoder}, $\varphi(X)$, that extracts the corresponding latent space for input $X$, and 2) the \textit{decoder}, $\zeta(\varphi(X))$,  that reconstructs a representation of the input image  from its compressed latest space representation. Ideally, the decoder can generate the exact inputs sample from the latent space, and the relation between the input and output of an AE can be expressed as $\zeta(\varphi(X)) = X $. However, in reality, the output of an AE is to some extent different from the input. This difference is known as reconstruction error and is defined as $E_R = |\zeta(\varphi(X)) - X|$ \cite{reconstruct1}. When training an AE, the objective is $E_R$.

A DAE is similar to AE, however, it is trained using a different training process.  As illustrated in Fig. \ref{ae}.b the input space of DAE are the noisy input samples, $X+\epsilon$, and their corresponding latent space is generated by $\varphi(X+\epsilon)$. Unlike AE (in which the $E_R$ is defined as the difference between the input and output of AE), the $E_R$ of DAE is defined as $E_R = |\zeta(\varphi(X+\epsilon)) - X|$ \cite{reconstruct1}. In other words, the reconstruction error is the difference between the output of decoder $\zeta(\varphi(X+\epsilon))$ and the clean input samples. An ideal DAE removes the noise $\epsilon$ from the noisy input and generates the clean sample $X$.

\begin{figure}[t]
  \centering
  \includegraphics[width=0.6\columnwidth]{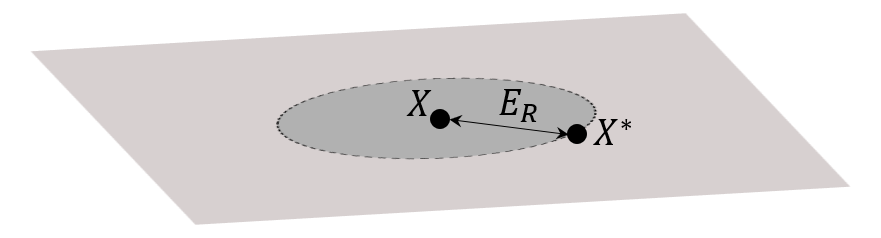}
  \caption {Reconstruction error ($E_R$) of a decoder can also result in mis-classification if the features extracted for the reconstructed image $X^*$ are pushed outside of the classifier's learnt decision boundary. }
 \label{adv}
\end{figure}

This refining property of DAEs, make them an appealing defense mechanism against adversarial examples. More precisely, by placing one or more DAEs at the input of a classifier, the added adversarial perturbations are removed and a refined input is fed into the subsequent classifier. The effectiveness of this approach highly depends on the extent of which the underlying DAE is close to an ideal DAE (in which the DAE completely refines the perturbed input). Although a well-trained DAE refines the perturbed input to some extent, it also imposes a reconstruction noises to it. As an example, assume that $\epsilon$ in Fig. \ref{ae}.b is zero.  This means the input $X$ is a clean image. In this case the output is $X+E_R$. If the size of $E_R$ is large enough, it can move the input $X$ over the classifier's decision boundary. This, as illustrated in Fig. \ref{adv}, will result in predicting the input $X$ as a $X^*$ class member. In this scenario, DAE not only fails to defend against adversarial examples, but also generates noise that could lead to the misclassification of the clean input images. 

The other problem of using AE or DAE as a pre-processing unit to refine the image and combat adversarial attacks is their added computational complexity. Adding an autoencoder as a pre-processor to a CNN increases 1) the energy consumed per classification, 2) the latency of each classification and 3the number of parameters of the overall model. 

In the following section, we propose a novel solution for protecting the model against adversarial attacks that addresses both the computational complexity problem and the reconstruction error issue of using an AE as a pre-processor.


\section{Proposed Method}\label{sec:method}

\begin{figure}[t]
  \centering
  \includegraphics[width=0.90\columnwidth]{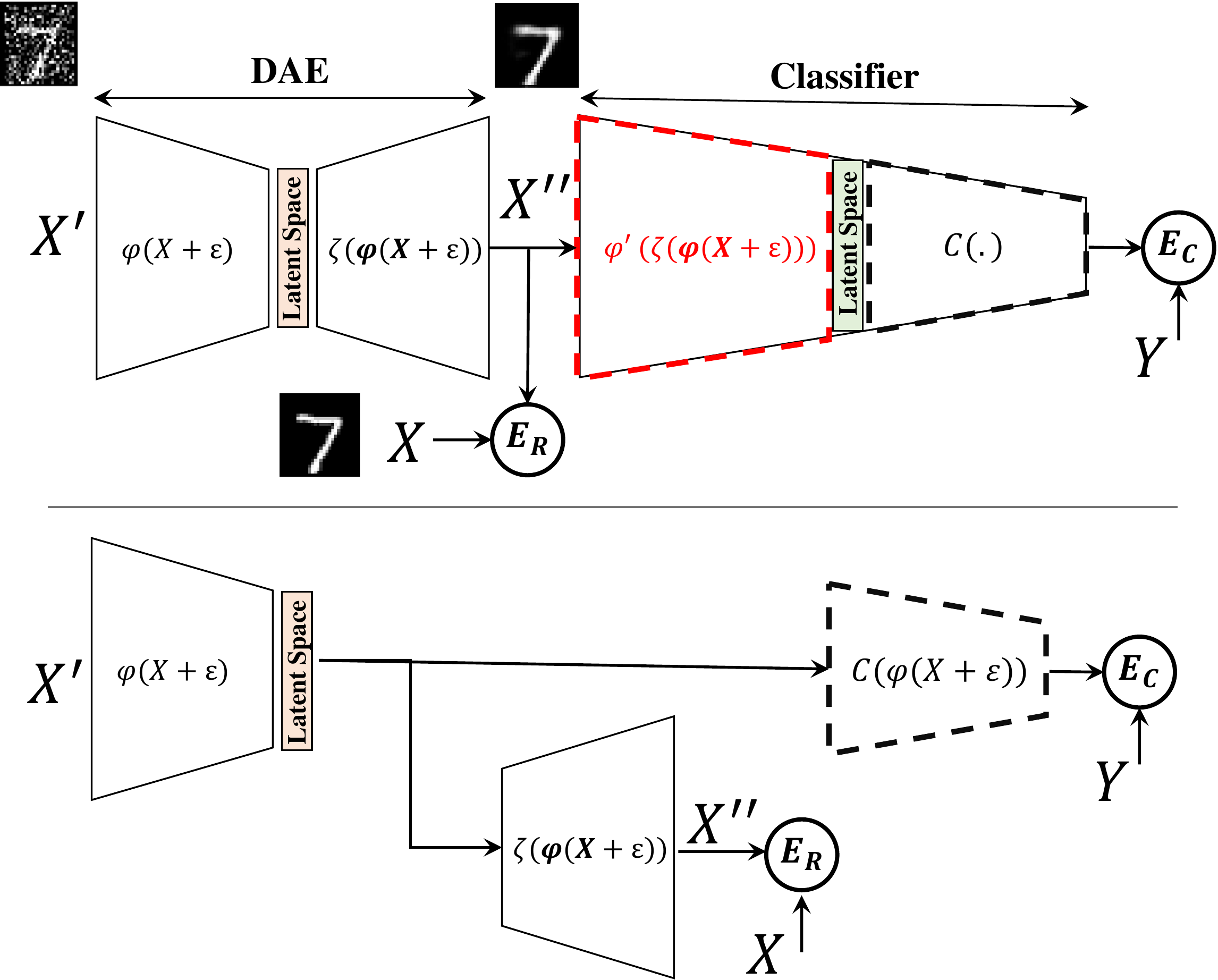}
  \caption {(Top) the defense proposed in \cite{magnet} where a DAE filter the noise in the input image before feeding it to a classifier. (Bottom) the CBC model in which the decoder of DAE and the first few conv layers of the base classifier are removed. Note that the decoder in CBC is only used for training the CBC, and is removed after training (for evaluation). In this figure $X$, $X'$, $X''$ are respectively the clean input sample, noisy input sample and the output of DAE. The $Y$ is the corresponding ground truth, and $E_R$ and $E_C$ are reconstruction error and classification error respectively. }
 \label{pdae}
\end{figure}

Using DAEs to refine perturbed input samples before feeding them into the classifier is a typical defense mechanism against adversarial examples \cite{comparative,magnet}. A general view of such defense is illustrated in Fig. \ref{pdae}.(top). In this figure, $\varphi(.)$, $\zeta(.)$ are the decoder and encoder of DAE, respectively, $\varphi'(.)$ represents the first few CONV layers of the CNN classifier, and $C(.)$, represents the later CONV stages. In this defense, the DAE and CNN are separately trained. The DAE is trained to minimize the reconstruction error, while the CNN is trained to reduce the pre-determined loss function (e.g. $L1$ or $L2$ loss). An improved version of such defense is when the training is done serially, where in the first stage, the DAE is trained, and then the CNN classifier is trained using the output of DAE as input sample. Note that the 2nd solution tends to get a higher classification accuracy. Regardless of the choice of training addition of a DAE to the CNN classifier adds to its complexity. Aside from added computational complexity, the problem with this defense mechanism is that AEs could act as a double agent: on one hand refining the adversarial examples is an effective means to remove the adversarial perturbation (noise) from the input image and is a valid defense mechanism, but on the other hand, its reconstruction error, $E_R$, could force misclassification of clean input images.  For correcting the behavior of the DAE, we propose the concept of Code Bridge Classifiers (CBC), aiming to  1) eliminating the impact of reconstruction error of the underlying DAE, and 2) reducing the computational complexity of the combined DAE and classifier to widen its applicability.


Fig. \ref{pdae}.(bottom), illustrates our proposed solution where the encoder $\varphi(.)$ of a trained DAE and a part of original CNN ($C(.)$) are combined to form a hybrid yet compressed model. In this model, the decoder $\zeta(.)$ of DAE, and the first few CONV layers of the CNN model, $\varphi'(.)$, are eliminated.  In CBC $\zeta(.)$ and  $\varphi'(.)$ are eliminated with the intuition that they act as an Auto Decoder (AD). As opposed to AE, the AD translates the latest space to an image and back to another latent space (intermediate representation of the image in the CNN captured by output channels of $\varphi'(.)$). This is, however, problematic because 1) the decoder $\zeta(.)$ is not ideal and it introduces reconstruction error to the refined image, 2) decoding and encoding (the first few CONV layers act as an encoder) of the image only translates the image from one latest space to another without adding any information to it. This is when such code translation (latest space to latent space) could be eliminated and the code at the output of $\varphi(.)$ could be directly used for classification. This allows us to eliminate the useless AD (the decoder $\zeta(.)$  and first few conv layers of original CNN that act as an encoder) and not only reduce the computational complexity the overall model but also improves the accuracy of the model by eliminating the noise related to image reconstruction of the decoder $\zeta(.)$. 

The training process for the CBC is serial: We first train a DAE, the encoder section of the model is separated. Then the trained decoder is paired with a smaller CNN compare to that of the original model. One way to build a smaller model is to remove the first few CONV layers of the original model and adjust the width of the DAE and the partial CNN to match the filter sizes. The rule of thumb for the elimination of the layers is to remove as many CONV layers equal to those in the encoder of AE. The next step is to train the partial CNN while fixing the values of the decoder, allowing the propagation to only alter the weights in the classifier $C(.)$.

\section{Implementation Details} \label{sec:framework}

In this section, we investigate the effectiveness of our proposed solution against adversarial examples prepared for FashionMNIST \cite{FashionMNIST} and CIFAR-10 \cite{cifar} datasets. To be able to compare our work with previous work in \cite{magnet}, we build our CBC solution on top of the CNN models that are described as \textit{Base} in tables \ref{tab:classifier1} and \ref{tab:classifier2}.  In these tables, the DAE columns represent the solution proposed in \cite{magnet}, in which a full auto-encoder pre-process the input to the CNN model and finally the columns CBC described the modified model corresponding to our proposed solution. 
The DAE as described in tables \ref{tab:classifier1} and \ref{tab:classifier2} includes 2 convolutional layers for encoding, and 2 convolutional transpose layers for decoding. The input is a clean image, and the output is an image of the same size generated by the autoencoder.

To build the CBC classifier, we stacked the trained encoder of the DAE with an altered version of the target classifier in which some of the CONV layers are removed. The trade-off on the number of layers to be removed is discussed in the next section. Considering that the encoder quickly reduces the size of the input image to a compressed latent space representation, the CNN following the latent space is not wide. For this reason, we also remove the max-pooling layers making sure that the number of parameters of the CBC classifier, when it reaches the softmax layer is equal to that of the base architecture. In our implementation, all the attacks and models are implemented using PyTorch \cite{pytorch} framework. To train the models we only use clean samples, and freeze the weights of the encoder part and train the remaining layers. Training parameters of the target classifier and the proposed architecture are listed in table \ref{tab:train}. We evaluated our proposed solutions against the FGSM \cite{fgsm}, Iterative \cite{bim}, DeepFool \cite{deepfool}, and Carlini Wagner \cite{cw} adversarial attacks. 

\begin{table}[t]
    \centering
  \caption{Architecture of the FashionMNIST Classifiers}
  \label{tab:classifier1}
 \scalebox{0.70}{
  \begin{tabular}{|l|l|l||l|l||l|l|}
    \hline
    \multicolumn{1}{|c|}{} &
    \multicolumn{2}{|c|}{Base} & 
    \multicolumn{2}{|c|}{DAE-CNN} &
    \multicolumn{2}{|c|}{CBC} \\
    \cline{2-7}
    & Type & Size & Type & Size  & Type & Size \\
    \hline
    \multirow{4}{*}{\rotatebox[origin=c]{90}{Defense}} & & & Conv.ReLU & $4 \times 4 \times 16$ & Conv.ReLU & $4 \times 4 \times 16$\\
    & & & Conv.ReLU & $4 \times 4 \times 48 $ & Conv.ReLU & $4 \times 4 \times 48$\\
    & & & ConvTran.ReLU & $4 \times 4 \times 48$ &  & \\
    & & & ConvTran.ReLU & $4 \times 4 \times 16$ &  & \\
    \hline
    \multirow{8}{*}{\rotatebox[origin=c]{90}{CNN}}&Conv.ReLU & $3 \times 3 \times 32$ & Conv.ReLU & $3 \times 3 \times 32$ & Conv.ReLU & $3 \times 3 \times 64$ \\
   & Conv.ReLU & $3 \times 3 \times 32$ & Conv.ReLU & $3 \times 3 \times 32$ & Conv.ReLU & $3 \times 3 \times 64$\\ 
   
  &Max Pool & $2 \times 2 $  &Max Pool & $2 \times 2$ & FC.ReLU & $4096 \times 200$\\
    &Conv.ReLU & $3 \times 3 \times 64$  &Conv.ReLU & $3 \times 3 \times 64$ &   FC.ReLU & $200 \times 200$\\
         &Conv.ReLU & $3 \times 3 \times 64$  &Conv.ReLU & $3 \times 3 \times 64$ &   Softmax & 10\\
    &FC.ReLU & $4096 \times 200$  &FC.ReLU & $ 4096 \times 200$ &   & \\ 
      &FC.ReLU& $200 \times 200$  &FC.ReLU & $200 \times 200$ &  & \\
  &Softmax & 10&Softmax & 10 & & \\
     \hline
\end{tabular}
}
\end{table}

\begin{table}[t]
    \centering
  \caption{Architecture of the CIFAR-10 Classifiers}
  \label{tab:classifier2}
 \scalebox{0.70}{
  \begin{tabular}{|l|l|l||l|l||l|l|}
    \hline
    \multicolumn{1}{|c|}{} &
    \multicolumn{2}{|c|}{Base} & 
    \multicolumn{2}{|c|}{DAE-CNN} &
    \multicolumn{2}{|c|}{CBC} \\
    \cline{2-7}
    &Type & Size & Type & Size  & Type & Size \\
    \hline
    \multirow{4}{*}{\rotatebox[origin=c]{90}{Defense}}& & &Conv.ReLU & $4 \times 4 \times 48$ & Conv.ReLU & $4 \times 4 \times 48$\\
    & & &Conv.ReLU & $4 \times 4 \times 72$ & Conv.ReLU & $4 \times 4 \times 72$\\
    & & &ConvTran.ReLU & $4 \times 4 \times 72$ &  & \\
    & & &ConvTran.ReLU & $4 \times 4 \times 48$ &  & \\
    \hline
    \multirow{13}{*}{\rotatebox[origin=c]{90}{CNN}}&Conv.ReLU & $3 \times 3 \times 96$ &Conv.ReLU & $3 \times 3 \times 96$ & Conv.ReLU & $3 \times 3 \times 96$ \\
   &Conv.ReLU & $3 \times 3 \times 96$ &Conv.ReLU & $3 \times 3 \times 96$ & Conv.ReLU & $3 \times 3 \times 192$\\ 
   &Conv.ReLU & $3 \times 3 \times 96$  &Conv.ReLU & $3 \times 3 \times 96$ & Conv.ReLU & $3 \times 3 \times 192$\\
  &Max Pool & $2 \times 2 $  & Max Pool & $2 \times 2$ & Conv.ReLU & $3 \times 3 \times 192$\\
    &Conv.ReLU & $3 \times 3 \times 192$  &Conv.ReLU & $3 \times 3 \times 192$ &   Conv.ReLU & $3 \times 3 \times 192$\\
         &Conv.ReLU & $3 \times 3 \times 192$  &Conv.ReLU & $3 \times 3 \times 192$ &   Conv.ReLU & $3 \times 3 \times 192$\\

    &Conv.ReLU & $3 \times 3 \times 192$  &Conv.ReLU & $3 \times 3 \times 192$ &   Conv.ReLU & $1 \times 1 \times 192$\\ 

      &Max Pool & $2 \times 2$  &Max Pool & $2 \times 2$ & Conv.ReLU & $1 \times 1 \times 192$\\

        &Conv.ReLU & $3 \times 3 \times 192$  &Conv.ReLU & $3 \times 3 \times 192$ &   Conv.ReLU & $1 \times 1 \times 192$\\

 & Conv.ReLU & $1 \times 1 \times 192$ & Conv.ReLU & $1 \times 1 \times 192$ & Avg Pool & \\
        & Conv.ReLU & $1 \times 1 \times 192$ &Conv.ReLU & $1 \times 1 \times 192$ & Softmax & 10 \\

    &Avg Pool & &Avg Pool & & & \\
    &Softmax & 10&Softmax & 10 & & \\
   \hline
          
\end{tabular}
}
\end{table}

\begin{table}[t]
    \centering
  \caption{Training Parameters}
  \label{tab:train}
  \scalebox{0.8}{
  \begin{tabular}{|l|l|l|l|l|l|}
  \hline
  Dataset & Optimization Method & Learning Rate & Batch Size & Epochs\\
  \hline
  FashionMNIST & Adam & 0.001 & 128 & 50\\
  CIFAR-10 & Adam & 0.0001 & 128 & 150
  \\

       \hline
           
\end{tabular}
}
\end{table}

\section{Experimental Results}\label{sec:results}

By adopting the training flow described in Table \ref{tab:parameter}, the top-1 accuracy of the base classifiers (in Tables \ref{tab:classifier1} and \ref{tab:classifier2}) that we trained for FashionMNIST and CIFAR10 are 95.1\%  90.\% respectively.  For the evaluation purpose, we trained denoising autoencoders with different noise values for both Datasets. The structure of the DAEs are shown in Table \ref{tab:classifier1} and \ref{tab:classifier2}. The reconstruction error for DAEs was arround 0.24 and 0.54 for FashionMNIST and CIFAR10 datasets respectively. 

\subsection{Selecting altered CNN architecture:}

\begin{figure}[t]
  \centering
  \includegraphics[width=\columnwidth]{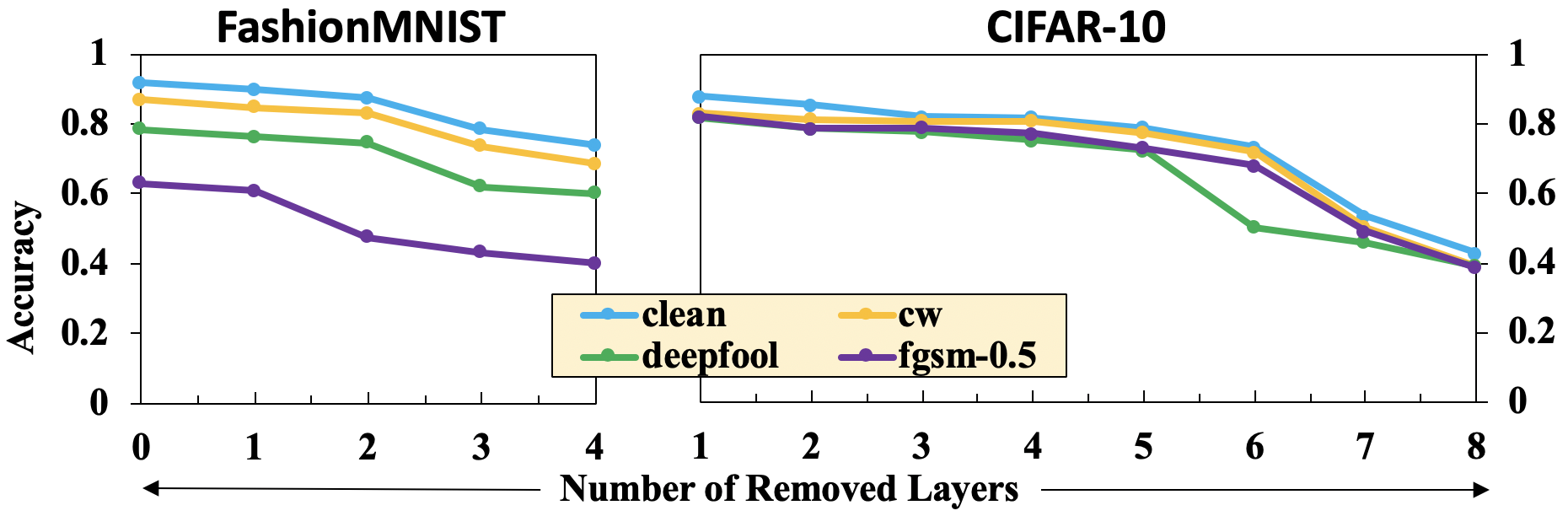}
  \caption{The change in the accuracy of the CBC model for a) FashionMNIST and b) CIFAR-10 classification with respect to the number of removed CONV layers from the base CNN model.}
  \label{fig:sw_layer}
\end{figure}

As discussed previously, the removal of the decoder of the DAE should be paired with removing the first few CONV layers from the base CNN and training the proceeding CONV layers to use the code (latent space) that is generated by the encoder as input. The number of layers to be removed was determined by a sweeping experiment in which the accuracy of the resulting model and its robustness against various attacks was assessed. Figure \ref{fig:sw_layer} shows the accuracy of CBC networks when the number of convolutional layers in the altered classifier is reduced compared to the base classifier. The experiment is repeated for both MNIST and CIFAR datasets, and the robustness of each model against CW \cite{cw}, Deepfool \cite{deepfool}, and FGSM \cite{fgsm} with $\epsilon=0.5$  is assessed.  As illustrated in Fig. \ref{fig:sw_layer}, the models remain insensitive to the removal of some of the first few layers (2 in MNIST, and 5 in CIFAR) with negligible (\~ 1\%) change in the accuracy by complete removal of each CONV layer until they reach a tipping point.  The MNIST model, for being a smaller model, reaches that tipping point when 2 CONV layers are removed, whereas the CIFAR model (for being a larger model) is slightly impacted even after 5 CONV layers are removed. 

\subsection{CBC accuracy and comparison with prior art}
\begin{figure*}[t]
  \centering
  \includegraphics[width=0.94\textwidth]{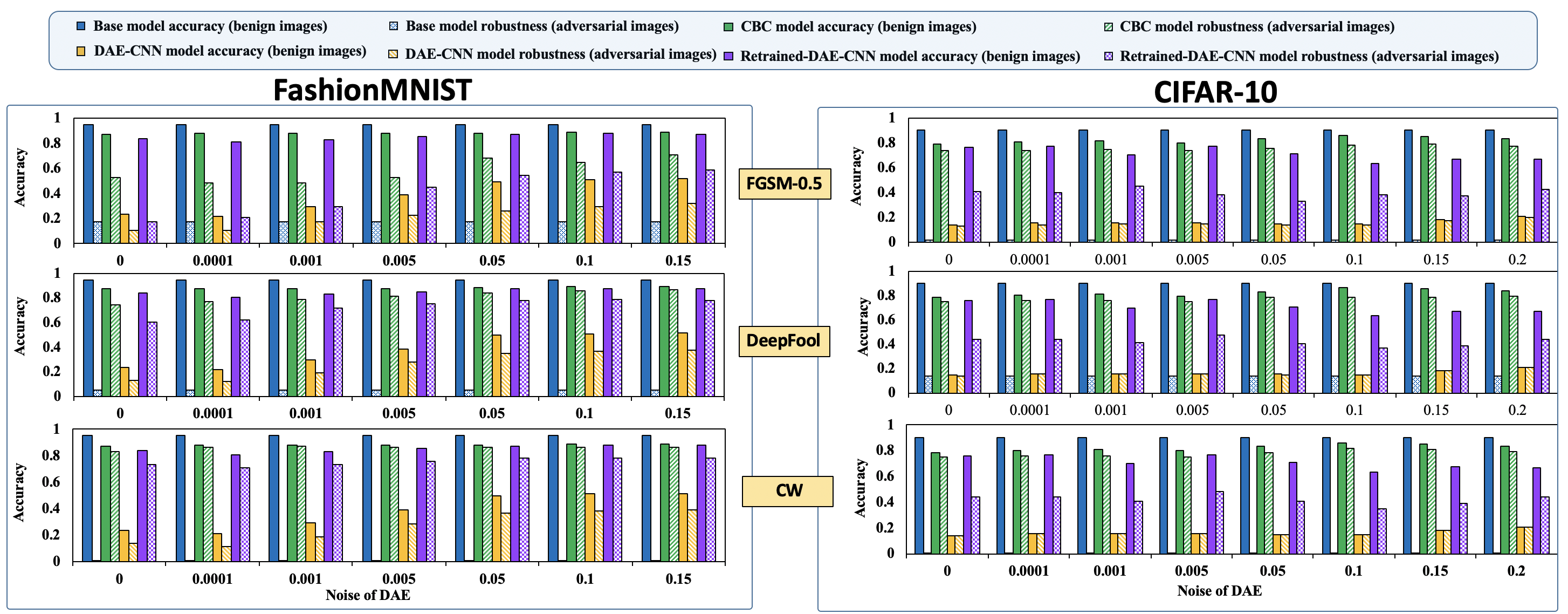}
  \caption{Comparing the accuracy of the base CNN model, the DAE protected CNN model (with and without retraining), and the CBC model when classifying bening images and adverserial images generated by different attack models: (left): FashionMNIST models, (right): CIFAR-10 models.}
  \label{fig:dae_all_results}
\end{figure*}

Fig. \ref{fig:dae_all_results} captures the result of our simulation, in which the robustness and the accuracy of the base CNN, the solution in \cite{magnet} in which the DAE refines the input for base CNN model, and our proposed CBC are compared. For the CNN protected with DAE, we provide two sets of results: 1) DAE-CNN Model accuracy: DAE and CNN are separately trained and paired together; 2) Retrained-DAC-CNN model accuracy: The CNN is incrementally trained using the refined images produced at the output of the DAE (denoted by Retraind-DAE-CNN). The comparison is done for the classification of both original and adversarial images. Results for  FGSM, DeepFool, and CW adversarial attacks are reported. For completeness, we have captured the robustness of each solution when the DAE is trained with different noise values.  


As illustrated in Fig. \ref{fig:dae_all_results}, the base model is very sensitive to adversarial examples, and the accuracy of the network in presence of adversarial examples (depending on the attack type) drops from over 90\% to the range of 0\% to  20\%.  The DAE-CNN model also performs very poorly even for benign images. This is because of the reconstruction error introduced by the decoder which severely affects the accuracy and the ability of the base CNN model. The Retrained-DAE-CNN model (representing the solution in \cite{magnet}) performs well in classifying benign images and also exhibits robustness against adversarial images. As illustrated, the robustness improves when it is paired with a DAE that is trained with high noise. The best solution, however, is the CBC solution:  regardless of the type of the attack, type of the benchmark, and the noise of DAE, the CBC model outperforms other solutions in both classification accuracy of the benign images and also robustness against adversarial examples. This clearly illustrates that the CBC model by eliminating the reconstruction error is a far more robust solution than DAE protected CNN models.  

\subsection{Reduction in model size and computational complexity}
In a CBC model, the DAE's decoder and the first few CONV layers of the base CNN model are removed. Hence, a CBC model has a significantly smaller flop count (computational complexity). Table  \ref{tab:parameter} captures the number of model Parameters and the Flop count for each of the CBC classifiers which are described in Tables \ref{tab:classifier1} and \ref{tab:classifier2}. Note that the majority of computation in a CNN model is related to its CONV layers, while a CONV layer has a small number of parameters. Hence, removing a few CONV layers may result in a small reduction in the number of parameters, but the reduction in the FLOP count of the CBC models is quite significant. As reported in Table \ref{tab:parameter}, in the FashionMNIST model, the flop count has reduced by 1.8x and 2.8X compared to the base and DAE protected model, while the parameter count is respectively reduced by 0.37\% and 2.69\% . This saving is more significant for the CIFAR-10 CBC model, where its computational complexity has reduced 3.1x and 3.3x compared to the Base and DAE protected model respectively, while the number of parameters is respectively reduced by 5.8\% and 13.4\%. Reduction in the flop count of the CBC model, as illustrated in table \ref{tab:parameter} also reduces the model's execution time. The execution time reported in table \ref{tab:parameter} is the execution time of each model over the validation set of each (FashionMNIST and CIFAR-10) dataset when the model is executed using Dell PowerEdge R720 with Intel Xeon E5-2670 (16 core CPUs) processors. As reported in table \ref{tab:parameter}, the execution time of the CBC is even less than the base CNN. Note that the CBC also results in processing unit energy reduction proportional to the reduction in the flop count. Hence, the CBC, not only resist against adversarial attacks, but (for being significantly smaller than the base model) also reduces the execution time, and energy consumed for classification.

\begin{table}[t]
  \centering
  \caption{Comparison of the number of parameters, computational complexity and execution time of CBC and the base model with AE and without AE protection.}

  \label{tab:parameter}
  \scalebox{0.88}{
  \begin{tabular}{|l|l|l|l|l|l|}
    \hline
  Dataset & Model & Flops & Parameters & Execution time \\
   \hline
   \multirow{3}{*}{FashionMNIST}& Base CNN & 9.08 MMac & 926.6 K & 463.4 s \\
   
    &AE-CNN\cite{magnet} & 14.3 MMac & 951.81 K  & 562.3 s\\
   
   &CBC & 5.04 MMac & 926.25 K & 293.7s \\
   \hline
   \multirow{3}{*}{CIFAR-10} & Base CNN& 0.59 GMac & 1.37 M &  1673.0 s\\
   
    &AE-CNN \cite{magnet} & 0.63 GMac & 1.49 M & 1749.7 s  \\
   
    &CBC & 0.19 GMac & 1.29 M & 1191.6 s \\
    \hline
           
\end{tabular}
}
\end{table}

\section{Conclusion}\label{sec:conclusion}

 In this paper, we propose the Code-Bridged Classifier (CBC) as a novel and extremely efficient mean of defense against adversarial learning attacks.  The resiliency and complexity reduction of CBC is the result of directly using the code generated by the encoder of a DAE for classification. For this purpose, at the training phase, a decoder is instantiated in parallel with the model to tune the denoising encoder by computing and back-propagating the image reconstruction error. At the same time, the code is used for classification using a lightweight classifier. Hence, the encoder is trained for both feature extraction (contributing to the depth of classifier and low-level feature extraction) and denoising.  The parallel decoder is then removed when the model is fully trained. This allows the CBC to achieve high accuracy by avoiding the reconstruction error of the DAE's decoder, while reducing the computational complexity of the overall model by eliminating the decoder and few CONV layers from the trained model.

\renewcommand{\IEEEbibitemsep}{0pt plus 0.5pt}
\makeatletter
\IEEEtriggercmd{\reset@font\normalfont\fontsize{7.0pt}{6.5pt}\selectfont}
\makeatother
\IEEEtriggeratref{1}

\bibliographystyle{IEEEtran}
\bibliography{IEEEabrv,Bibliography}

\end{document}